\documentclass{article}

\usepackage{microtype}
\usepackage{graphicx} 
\usepackage{booktabs} 
\usepackage{subcaption}
\usepackage{hyperref}
\usepackage{amsmath,amssymb,amsfonts}

\usepackage[accepted]{icml2021}

\icmltitlerunning{Adversarial Stacked Auto-Encoders for Fair Representation Learning}

\begin{document}

\twocolumn[
\icmltitle{Adversarial Stacked Auto-Encoders for Fair Representation Learning}



\icmlsetsymbol{equal}{*}

\begin{icmlauthorlist}
\icmlauthor{Patrik Joslin Kenfack}{to}
\icmlauthor{Adil Mehmood Khan}{goo}
\icmlauthor{Rasheed Hussain}{to} 
\icmlauthor{S.M. Ahsan Kazmi}{to} 
\end{icmlauthorlist}

\icmlaffiliation{to}{Networks and Blockchain Lab, Innopolis University, Innopolis, Russia.}
\icmlaffiliation{goo}{Machine Learning and Knowledge Representation Lab, Innopolis University, Innopolis, Russia}  

\icmlcorrespondingauthor{Patrik Joslin Kenfack}{p.kenfack@innopolis.university}
\icmlcorrespondingauthor{Rasheed Hussain}{r.hussain@innopolis.ru}

\icmlkeywords{Machine Learning, ICML}

\vskip 0.3in
]



\printAffiliationsAndNotice{} 

\begin{abstract}

Training machine learning models with the only accuracy as a final goal may promote prejudices and discriminatory behaviors embedded in the data. One solution is to learn latent representations that fulfill specific fairness metrics. Different types of learning methods are employed to map data into the fair representational space. The main purpose is to learn a latent representation of data that scores well on a fairness metric while maintaining the  usability for the downstream task. 
In this paper, we propose a new fair representation learning approach that leverages different levels of representation of data to tighten the fairness bounds of the learned representation. Our results show that stacking different auto-encoders and enforcing fairness at different latent spaces result in an improvement of fairness compared to other existing approaches. 
\end{abstract}

\section{Introduction}
\label{submission}
Representation learning has made a significant mark in the field of Machine Learning (ML) over the past decade. It has technologies that extract useful information or features from data to improve the classification or predictive performance of models, or even generate new synthetic realistic data. Several applications for different kind of tasks have emerged such as, machine translations \cite{baltruvsaitis2018multimodal}, anomalies detection \cite{rivera2020anomaly}, objects and actions recognition \cite{papageorgiou2000trainable}, etc.

ML models are widely used in real life to make decisions that can affect people's lives, e.g., loan applicant, college admission, criminal justice, hiring, etc. 
Models trained with biased data can lead to unfair decisions \cite{mehrabi2019survey}. In fact, these models mainly rely on human-generated data to learn patterns that are then used to make predictions on the new unseen data. However, real-world data are already tainted by prejudices and unfair decisions (historical bias), which reflect the flaws of our society. Historical bias is one origin of algorithmic bias. Another source of algorithmic bias is the representation bias \cite{mehrabi2019survey}. It arises when certain groups of the population are underrepresented within the data. For example, a facial recognition model trained with data containing considerably more white faces than black faces will tend to be less accurate when used on black faces.
To this end, the algorithmic bias occurs when biases in the data are learned by the model and therefore lead to unfair decisions \cite{dwork2012fairness, joslin2021fairness, hardt2016equality}. 

One approach to mitigate the impact of biases from the data is the {\it fair representation learning}. With this technique, the input data is mapped into a new representation which is enforced to satisfy a given fairness metric while maintaining the utility of the representation as much as possible. The learned representation can then be used for any downstream task such as classification or data generation, with better chances of yielding fair results.  
Existing works by ~\citeauthor{madras2018learning, edwards2015censoring} used adversarial learning to enforce the fairness of the representation with respect to statistical parity. They used an auto-encoder as a generator whose aim is to learn a latent space such that an adversary cannot predict the sensitive feature (gender, race, etc.) from the learned latent representation. In \citeauthor{madras2018learning}, the authors proposed a learning objective for other fairness metrics such as equalized odds and equal opportunity (section \ref{sec:related}) with theoretical bounds of fairness. 

This work builds on top of the previous works where we propose a fair representation learning approach based on adversarial stacked auto-encoders. However, our proposed approach leverages different level of representation of the input data to tighten the fairness bounds of the learned representation. In fact, the success of deep networks can be attributed to their ability to exploit the unknown structure in the input distribution to discover useful features at multiple levels.  In this multi-level representations, the higher-level learned features are defined in terms of lower-level features ~\cite{bengio2013representation}. For instance, Khan et al.~\citeauthor{khan2020post} showed that performing data augmentation in the feature space (and at different level of representation), can improve predictive performances of the neural network. Similarly, a generative model proposed by Huang et al. ~\citeauthor{huang2017stacked} leveraged different level of representation to improve the quality of the generated images. Applying fairness at a given level does not guarantee that information about the sensitive attribute is removed, as it may not all be presented at the given level.

In essence, we hypothesize that the above arguments may also be useful for improving fairness, which was confirmed by our empirical results. Intuitively, the main idea is to approach an optimal adversary via sequential learning in which one adversary is used to enforce fairness on a low-level representation. This low-level representation is then used as input for a higher-level representation on which another adversary is trained to enforce fairness on that representation by improving the previous adversary.       

The reminder of the paper is organized as follow. In Sections \ref{sec:related} and \ref{sec:background}, we present related work and background, respectively. In Section \ref{sec:method}, we introduce our fair representation learning approach that tightens the fairness bounds. In Section \ref{sec:result}, we present empirical results which show the effectiveness of our learned representation on several real-life datasets. In Section \ref{sec:conclusion}, we conclude the paper.

\section{Related Work}
\label{sec:related}

Pre-processing techniques are used to mitigate biases from the data by enforcing a given fairness property while maintaining the utility of the predictions. The objective of fair representation learning is to learn a representation of the data that is most likely to produce fair results for downstream tasks. In \cite{zemel2013learning}, the authors presented the first fair representation learning approach which removes dependencies on the sensitive attribute by mapping input data to new points called prototypes.   Prior work in this direction focuses on statistical parity, equalized odds, and equal opportunities.  

The goal is to learn a representation that will remove all the dependencies in regards to the sensitive attribute from the training data, while retaining as much information as possible. In \cite{louizos2015variational}, the authors proposed the Variational Fair Auto-Encoder (VFAE), a variant of variational auto-encoder that maps the input data into a latent space while discarding information about the sensitive attributes from the data as much as possible. Thus the sensitive attributes are treated as nuisance variable. To do this, the authors (i) used a factorized prior $ p(z)p(s) $ where $z$ is the latent representation and $s$ is the sensitive attribute, and (ii) added a regularization term to encourage the independence between $z$ and $s$ using maximum mean discrepancy. 
%

In \cite{edwards2015censoring}, the authors proposed an approach to learn fair representation using adversarial learning to enforce demographic parity. Similarly, in \citeauthor{beutel2017data}, the authors explored the particular fairness levels achieved by the algorithm from \cite{edwards2015censoring} and showed how other fairness metrics can be achieve by varying the distribution of the adversary's input. Madras et al. \citeauthor{madras2018learning} extended the previous work by proposing adversarial objectives that yield fair and transferable representations that in turn admit fair classification outcomes. They provided adversarial objective functions for each fairness metric that upper bounds the unfairness of arbitrary downstream classifiers in the limit of adversarial training.

In this work, we propose a new fair representation learning approach built upon previous works, which aims to improve the fairness of models through stacked adversarial learning. We enforce fairness at different levels of representation in order to tighten the fairness bounds of the final representation.    

\section{Background}
\label{sec:background}
In this section, we introduce fairness notions used throughout this paper and concepts related to fair representation learning and adversarial learning.
\subsection{Fairness}
Consider a training data $\mathcal{D}=\{X, Y, S\}$, where $x_i \in \mathbb{R}^n$ is the feature vector, $y_i \in \{0, 1\}$ is the label, and $S$ is the binary protected attribute (e.g., gender, race, etc.). Learning a fair representation means mapping the input data $X$ into a new representation $X'$ such that $X'$ will satisfy one of the following fairness criteria:

\begin{itemize}
	\item \textit{Statistical parity}: It is also known as Demography parity ($\Delta_{DP}$). This fairness criteria promotes the independence between the predictor outcome ($\hat{Y}$ a function of $X'$) and the sensitive attribute. $\hat{Y} \bot \; S$, i.e., a predictor  satisfies the statistical parity if  $P(\hat{Y}|S=0) = P(\hat{Y}|S=1)$ \cite{dwork2012fairness}. For example, a loan approval system will achieve statistical parity if it does not deny loans to men more often than to women. A drawback of this fairness criterion is that it allows unqualified applicants to be selected as long as the acceptance rate is the same for both groups.  However, when the sensitive attribute correlates with the target variable, a drop in accuracy can be observed.   
	
	\item \textit{Equalized Odds}: In contrast to $\Delta_{DP}$, Equalized Odds ($EO$) promotes the conditional independence between the prediction outcome and the sensitive attribute given the class label ($\hat{Y} \bot S| Y$). A predictor outcome $\hat{Y}$ trained with $X'$ satisfies EO if  $P(\hat{Y}=y|S=0, Y=y) = P(\hat{Y} = y|S=1, Y=y), \forall y \in \{0, 1\}$. In other words the False Positive Rate (FPR) and the True Positive Rate (TPR) of groups should be the same. One advantage of equalized odds is that it admits the perfect model $\hat{Y} = Y$ \cite{hardt2016equality, verma2018fairness}. 
	\item \textit{Equal opportunity}: Similar to EO, Equal opportunity (EOpp) only considers the case where $Y=1$ ($\hat{Y} \bot S| Y=1$). 
	A predictor outcome $\hat{Y}$ satisfies EOpp if $P(\hat{Y}=1|S=0, Y=1) = P(\hat{Y} = 1|S=1, Y=1)$. In other words, groups should have the same TPR.
\end{itemize}

It is worth noting that the predictor trained with fairness constraints are less accurate than the ones trained without it \cite{kamishima2011fairness}. Thus fairness comes at the expense of accuracy. A desired property is to provide fair representation with lower fairness accuracy trade-off.

\subsection{Adversarial Learning}
Inspired by the game theory, adversarial learning consists of two neural networks (generator and discriminator) trained in an adversarial manner. The generator's ($G$) goal is to fool the discriminator ($D$) by sampling as realistic examples as possible such the discriminator --which the goal is to distinguish between fake samples and real samples--, will not be able to make the difference between examples $G(z)$ sampled from $G$ using the random noise vector $z$ and real examples $x$. Thus, $G$ and $D$ play a min-max game with value function $V(G, D)$:

\begin{equation} 
\begin{split}
\begin{aligned}
    \underset{G}{\mathrm{min}} \: \underset{D}{\mathrm{max} }\: V(D, G) = \; & \mathbb{E}_{x \sim p_{data}(x)}[\mathrm{log} \: D(x)]\\ 
    &+ \mathbb{E}_{z \sim p_{z}}[1 - \mathrm{log}\: D(G(z))]
\end{aligned}
\end{split}
\end{equation}
where $D$ seeks to maximize this quantity while $G$ seeks to minimize it.

\section{Methods}
\label{sec:method} 
\begin{figure*}[htbp]
	\centering
	\includegraphics[width=\textwidth]{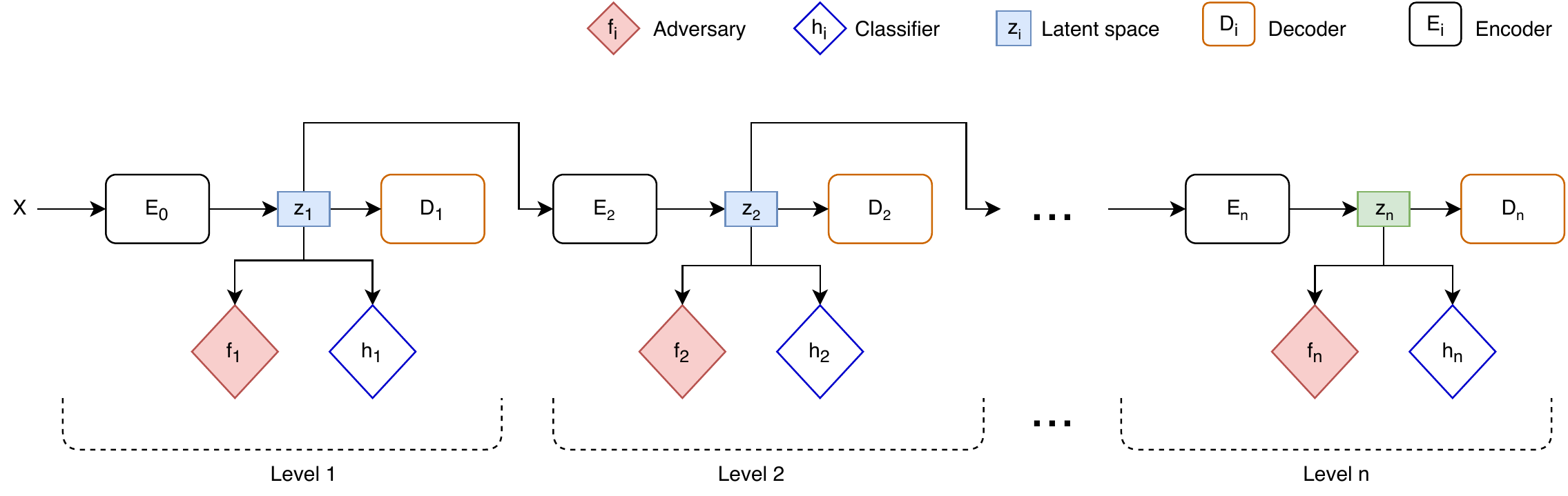}
	\caption{Adversarial Stacked Auto-Encoders architecture}
	\label{fig:archi}
\end{figure*}
In this section, we describe the architecture of our proposed model and the training procedure. Figure \ref{fig:archi} presents an overview of the architecture and the training process.   

\subsection{Model Architecture}
Our main idea is to stack different Encoders ($E_i$), Decoders ($D_i$), classifier $f_i$, and adversary ($h_i$), in order to get different levels of representation of the input data.
The intuition here is that, different level of representation can exhibit different details of information from the data. Enforcing fairness at a given level does not guarantee that fairness bounds are tight enough, unless the adversary is an optimal one, which may not be available in non-convex settings. Our goal is to approach this optimal adversary in an incremental may. 
   
At a each level $i$, we have different components: the learned representation $z_i$  yielded by the encoder $E_i$, the corresponding decoded representation $z'_i$ produced by the decoder $D_i$, the adversary network $f_i$ that enforces the fairness of that representation and the predictor network $h_i$ that enforces the utility of the representation. $z_0$ represents the input data $X$, and $z'_0$ the final reconstructed output ($X'$). 
The overall loss at each level $i$ is defined as the linear combination of three loss terms: the reconstruction loss ($\mathcal{L}^{rec}_{E_i, D_i}$),  the adversary loss ($\mathcal{L}^{adv}_{f_i}$) and the predictor loss ($\mathcal{L}^{adv}_{h_i}$):
\begin{equation}
\mathcal{L}(G_i, D_i, f_i, h_i) = \alpha \mathcal{L}^{rec}_{E_i, D_i} + \beta\mathcal{L}^{Adv}_{f_i} + \gamma\mathcal{L}^{Class}_{h_i}
\end{equation}

In the above equation, $\alpha$, $\beta$ and $\gamma$ are the weights associated with each loss.
Thus, $\mathcal{L}^{rec}_{E_i, D_i}$ is the loss of reconstructing the encoded representation $z_{i}$ by the decoder $D_i$. For the reconstruction loss we use the Root Mean Squared Error (RMSE):
$\mathcal{L}^{rec}_{E_i, D_i} = \frac{1}{|X|} ||z'_{i} - D_i(E_i(z_{i-1}))||^2_2$. The adversarial loss is to enforce the representation to satisfy certain fairness constraint. For instance, to satisfy statistical parity, the adversary loss is defined as cross entropy loss:
\begin{equation}
\mathcal{L}^{adv}_{f_i} = \frac{1}{|X|} \sum_{s, \hat{s} \in S, \hat{S}} s \cdot \mathrm{log}(\hat{s} + (1-s) \cdot \mathrm{log}(1-\hat{s})) 
\end{equation}

The adversary network at the level $i$ tries to minimize the loss of predicting the sensitive attribute $S$ from the encoded representation $z_i$, while the predictor and the generator (typically auto encoder) try to maximize it. The losses of predictor and adversary can be defined as cross entropy loss or using loss functions proposed in \cite{madras2018learning} to satisfy equalized odds and equal opportunities.
Thus at each level, we have the following min-max problem: 

\begin{equation}
\label{eq:minimax}
\underset{G_i, D_i, h_i}{\mathrm{min}} \; \underset{f_i}{\mathrm{max}} \; \mathcal{L}(G_i, D_i, f_i, h_i)
\end{equation}

To have a different representation at each level, we vary the dimension of each latent space, from higher to lower dimensions ($|z_i| > |z_{i+1}|$).   

\subsection{Model training}
At a given level $i$, we realize the classifier, auto-encoders and adversary as neural networks and alternate gradient decent and ascent steps to optimize their parameters according to \ref{eq:minimax}. First the encoder-classifier-decoder takes a gradient step to minimize $\mathcal{L}$ (Equation \ref{eq:minimax}) while the adversary $f_i$ is fixed, then $f_i$ takes a step to maximize $\mathcal{L}$ with fixed auto encoder and classifier. We use a relaxation of adversary objectives proposed by \cite{madras2018learning, beutel2017data}, i.e., to achieve Equalized Odds, in addition to the latent space $z$,  we passed the class label $Y$ to the adversary. To achieve Equal Opportunity, the loss function (Eq \ref{eq:minimax}) is computed only using samples where $Y=0$.

The training is performed sequentially, starting with an initial latent representation $z_1$ trained using the input data. During the first training, the adversary $f_1$ enforces fairness (typically $\Delta_{DP}$, $\Delta_{EO}$, or $\Delta_{EOpp}$) of the lower level representation $z_1$. Afterwards, a new latent space of lower dimension $z_2$ (higher level representation) is stacked, and uses the pre-trained representation $z_1$ as input. 

The number of stacked layers on which the fairness constraints are imposed depends on the depth of the neural network and are specified as a hyper parameter. In the experiments, we used a Multi Layers Perceptron (MLP) network for the encoder and decoder with one hidden layer. Initially, fairness is applied on the hidden layer ($z_1$), then the output layer (latent space) is stacked and used as the final representation ($z_2$). In the testing phase, we get rid of all the decoders, adversaries, and classifiers. Only the encoders are used to map the input data into the fair space.    

\section{Experiments}
\label{sec:result}

\begin{figure*}[t]
    \centering
        \begin{subfigure}{0.33\textwidth}
     \includegraphics[width=\linewidth]{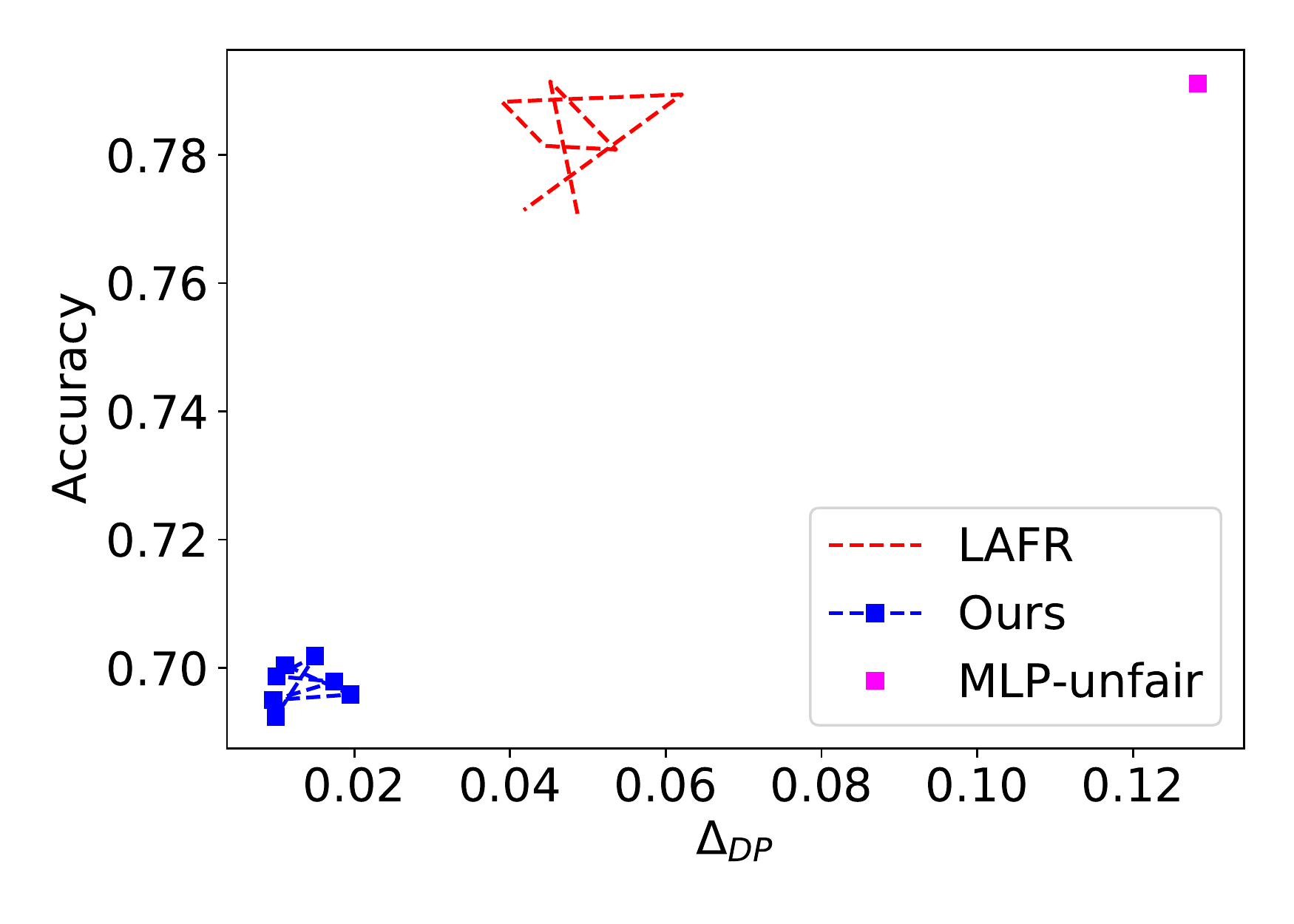}
     \caption{German Dataset}
     \label{fig:blogs-2dom-acc}
    \end{subfigure}
    \begin{subfigure}{0.33\textwidth}
     \includegraphics[width=\linewidth]{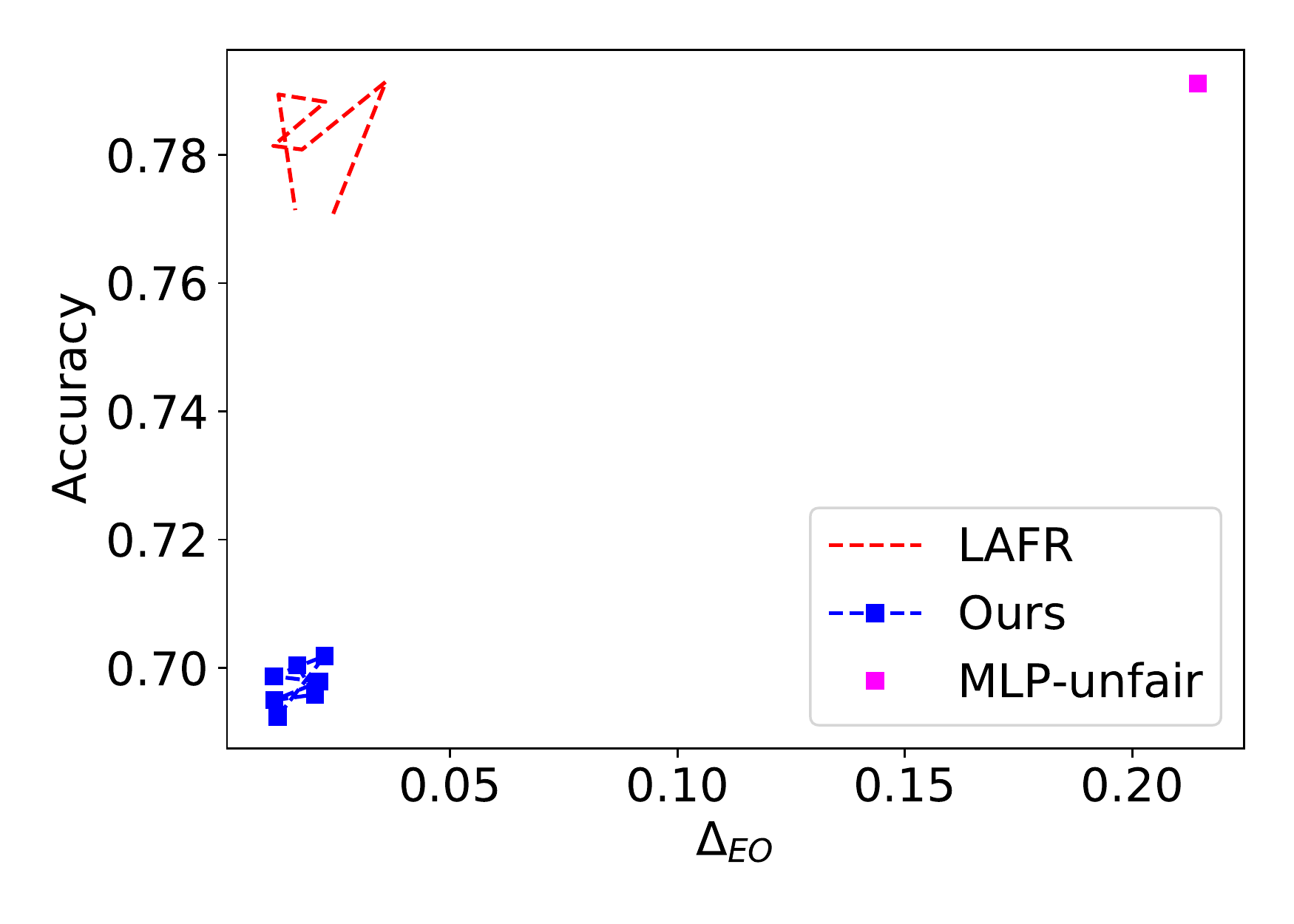}
     \label{fig:blogs-2dom-gpt}
    \end{subfigure}
    \begin{subfigure}{0.33\textwidth}
     \includegraphics[width=\linewidth]{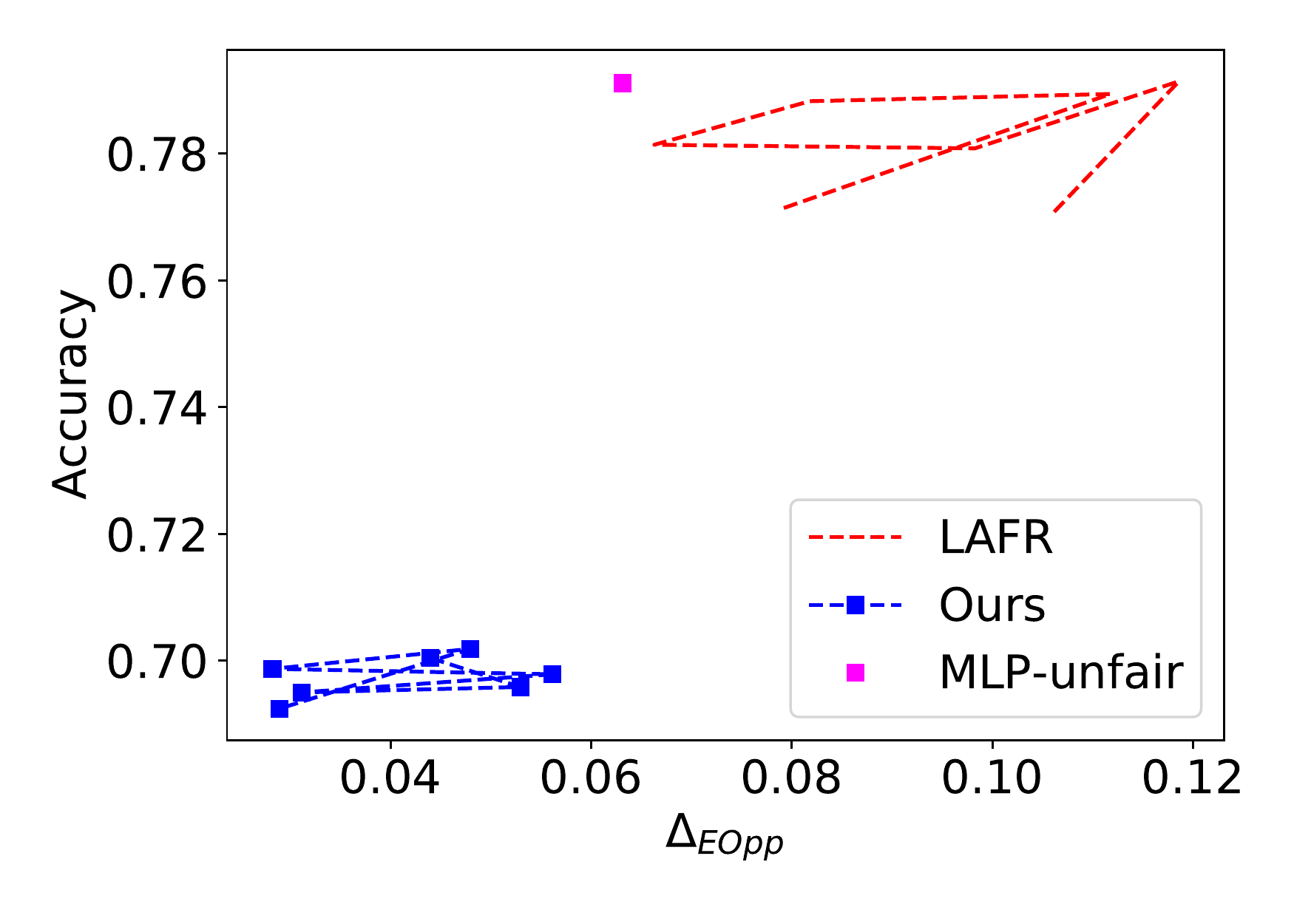}
     \label{fig:blogs-2dom-unc}
    \end{subfigure}
    
    \begin{subfigure}{0.33\textwidth}
     \includegraphics[width=\linewidth]{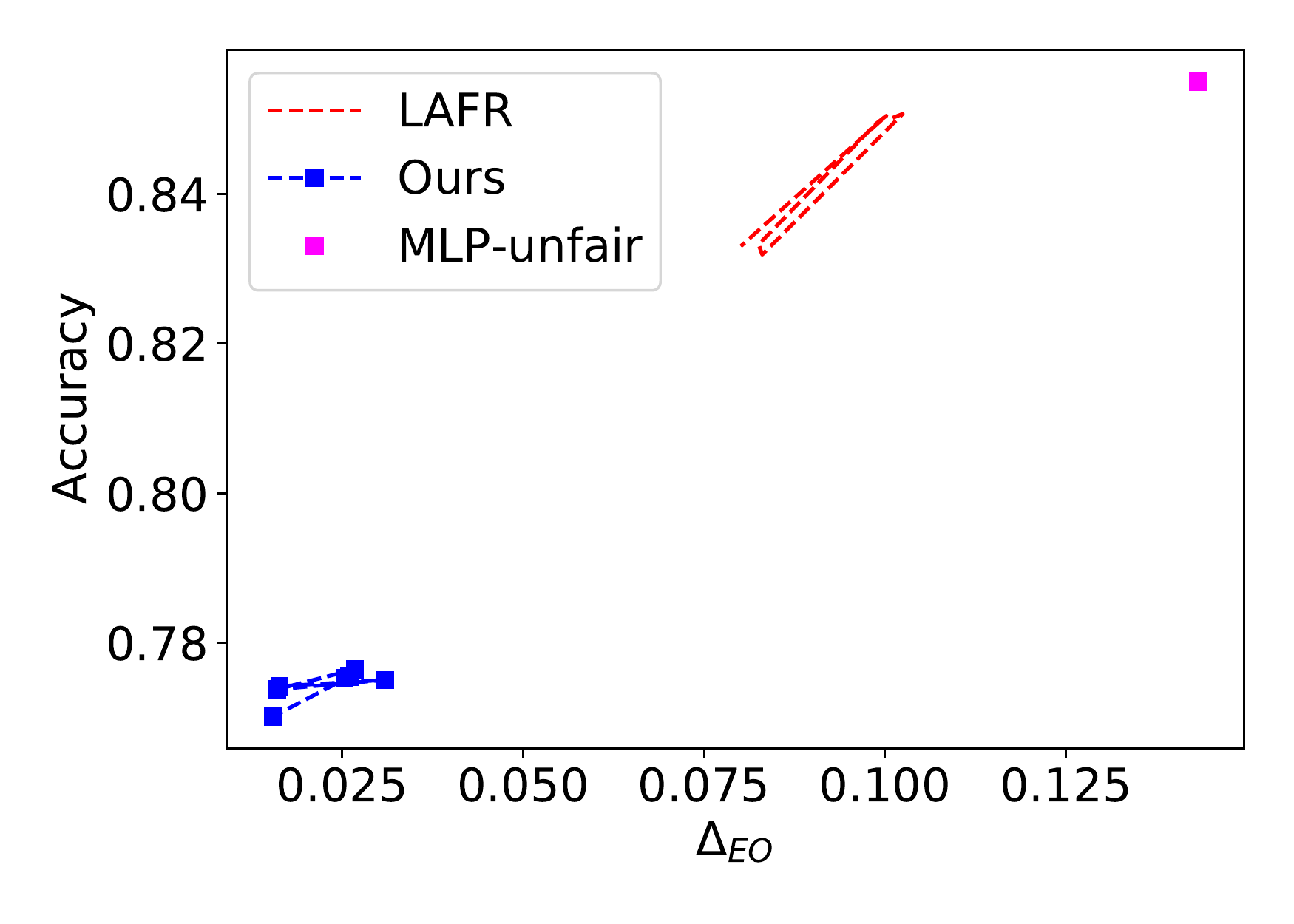}
     \caption{Adult Dataset}
     \label{fig:blogs-2dom-acc}
    \end{subfigure}
    \begin{subfigure}{0.33\textwidth}
     \includegraphics[width=\linewidth]{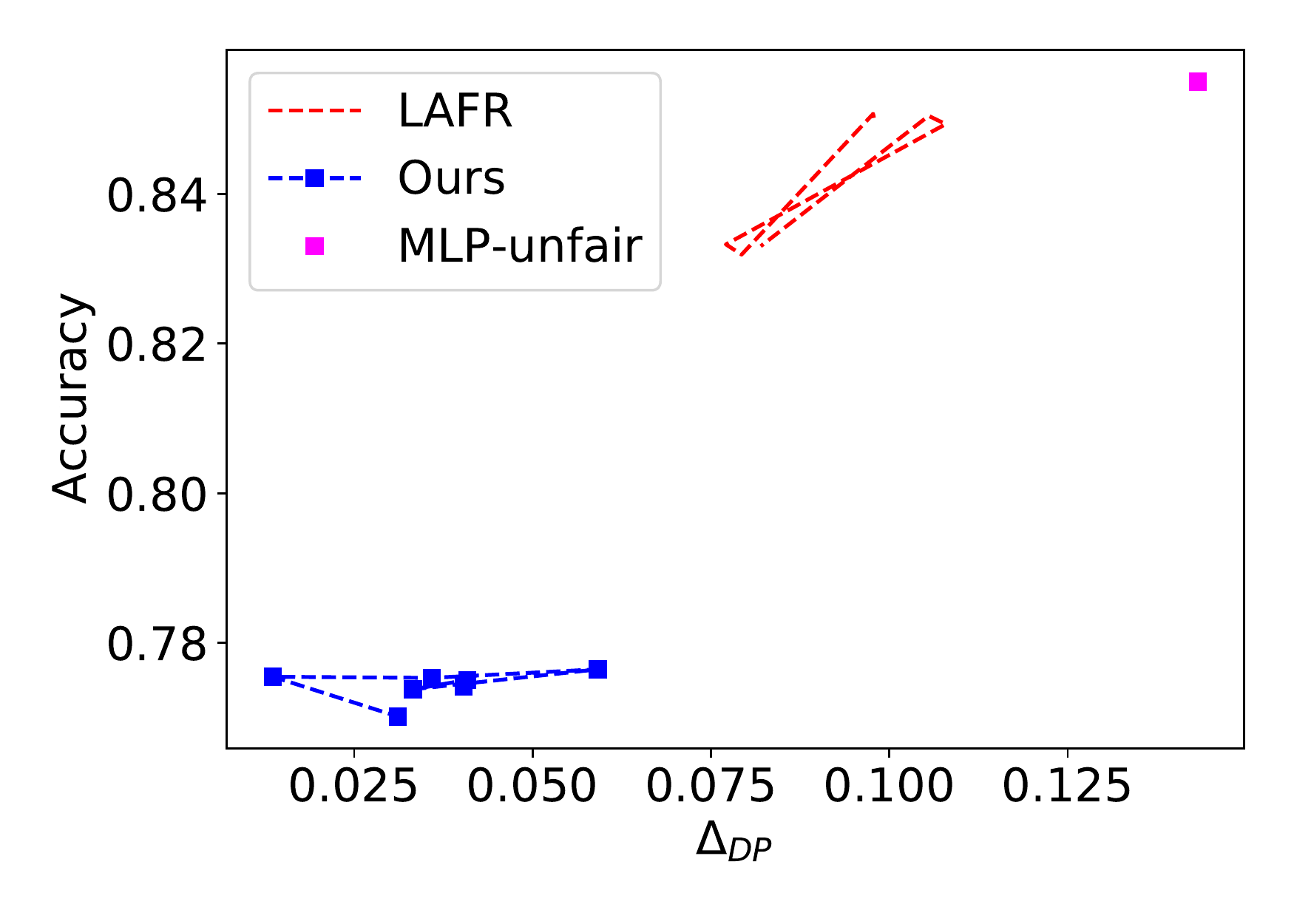}
     \label{fig:blogs-2dom-gpt}
    \end{subfigure}
    \begin{subfigure}{0.33\textwidth}
     \includegraphics[width=\linewidth]{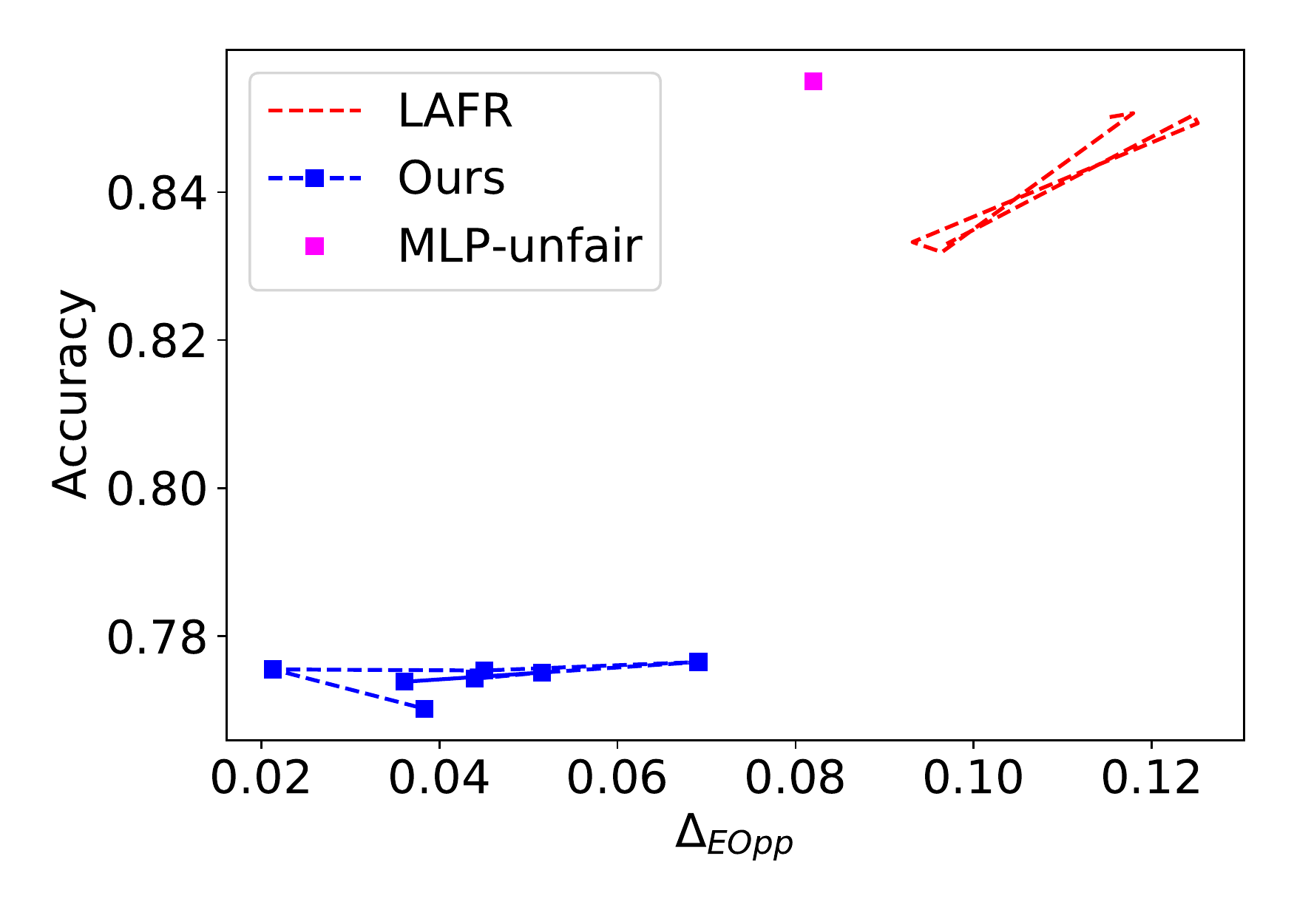}
     \label{fig:blogs-2dom-unc}
    \end{subfigure}
    \caption{Accuracy-fairness trade-offs of classification tasks on the German (first row) and Adult (second rows) datasets. Our learned representation always lower bounds the fairness results of the representation learned by vanilla approach (LAFR). It shows that the fairness bounds of our approach is more tight. However, we can observe a slight decrease in accuracy compared to other representations. }
    \label{fig:fair-classification}
    \vspace{-2ex}
\end{figure*} 

We present experiments on two standard real world datasets widely used for fair classification as suitable  benchmarks to  compare  the  performance  of  different  machine  learning methods: 
The  Adult Income dataset  \cite{asuncion2007uci}  has  48843  instances  of  demographic  information  of  American  adults,  described  with  14 features.  The  target  variable  indicates  whether  individual’s income is larger than 50K US dollars. 
The German credit dataset  \cite{compas} has 1000 instances of bank account  information  represented  by  20  features  with  the  aim to classify bank account holders into credit class good or bad. For both datasets, we use gender as the single protected (sensitive) attribute. We demonstrate the effectiveness of our approach compared to standard fair representation learning techniques. The standard (vanilla) approach is fair representation learning with fairness applied only to one latent space.

\subsection{Fair Classification}
Figure \ref{fig:fair-classification} shows the fairness results of the MLP trained with the representation obtained by our approach compared to the representation produced by the vanilla approach (Learning Adversarial Fair Representation $-$ LAFR) and MLP trained with original input data (MLP-unfair). For the vanilla approach, we used a network architecture with one hidden layer of 20 units, and latent space of 8 units for Adult dataset, 15 hidden units and 8 output units for the German dataset. We trained models with the same architecture using LAFR approach and our approach with two levels of representation, i.e., we trained an adversary on the hidden layer and then stacked the output layer and trained another adversary on it. We used single-hidden-layer neural networks for each of our classifier and adversary with 20 hidden units. The equation \ref{eq:minimax} is optimized using Adam optimizer \cite{kingma2014adam} with learning rate of $0.01$, a batch size of 64, trained for 150 epochs for Adult dataset and 1000 for the German credit. We run the experiment seven times with different values of $\beta$ $(1, 2, 3, 5, 15)$, with  $\alpha = 0$ and $\gamma = 1$. 

Similar to the process used by \citeauthor{madras2018learning}, we created a feed-forward model which consisted of our frozen, adversarially-learned encoders followed by an MLP with one hidden layer, with a loss function of cross entropy with no fairness modifications. We reported the mean over all runs per $\beta$ and we used a validation procedure to evaluate. The results shows that representation produced by our model always lower bounds the fairness of standard approaches. This shows that our approach provides tighter fairness bounds. However, since the main objective of our work is to improve the fairness, a decrease in accuracy is observed compared to the standard approach, which we attribute to the trade-off between fairness and accuracy.

\subsection{Classification on Downstream Tasks}

\begin{table}[htbp] 
\caption{Comparison of $\Delta_{DP}$ on classification tasks using  logistic regression and random forest model on Adult and German datasets}
\label{tab:results}
\vskip 0.15in
\begin{center}
\begin{small}
\begin{sc}
\resizebox{\columnwidth}{!}{%
\begin{tabular}{llll}
\toprule
Model & Unfair & LAFR & Ours \\
\midrule 
\multicolumn{4}{c}{Adult}  \\
\hline
Logistic Regression & 0.53$\pm$ 0.008  & 0.51$\pm$ 0.009 & \textbf{0.21}$\pm$ 0.004\\
Random Forest &  0.54$\pm$ 0.001 & 0.49$\pm$ 0.001& \textbf{0.25}$\pm$ 0.007\\ 
\hline
\multicolumn{4}{c}{German}  \\
\hline
Logistic Regression & 0.36$\pm$ 0.08  & 0.33$\pm$ 0.09 & \textbf{0.08}$\pm$ 0.04\\
Random Forest &   0.27$\pm$ 0.03  &  0.23$\pm$ 0.06 & \textbf{0.11}$\pm$ 0.05\\ 
\bottomrule
\end{tabular}%
}
\end{sc}
\end{small}
\end{center}
\vskip -0.1in
\end{table}

Learning fair representation is a model-agnostic approach to mitigate unfairness, i.e., the learned representation can be used for any downstream task and not only for neural network based models. We tested linear and non-linear models on representation produced by our model and standard approach. We trained the representation using the network architecture described in previous section, without hyper-parameter tuning and using $\alpha = 0$, $\beta = 1$, $\gamma = 1$. We also trained models on the original dataset without fairness constraints.

Table \ref{tab:results} shows $\Delta_{DP}$ reported from 5-fold cross validations on Adult and German datasets. Results shows that the representation produced by our model also provides better fairness performances when trained using classical machine leaning algorithms such as Linear Regression and Random Forest. We observed similar results for other fairness metrics (EO, EOpp).



\section{Conclusion}
\label{sec:conclusion}
In this paper, we showed that applying fairness at different levels of representation improves the fairness performance of the learned representation. In this regard, we proposed an adversarial stacked auto-encoder architecture which expose different level of representation of the input data, on which several adversary networks are trained sequentially to tighten the fairness bounds of the final representation (lowest level representation). 

Our empirical results show that this approach outperforms standard adversarial fair representation learning approach in terms of fairness. Intuitively, our learning process lead to learning an optimal adversary in incremental way. However, stabilizing adversarial training of fair representations remains an important issue that we plan to address in our future work.







\nocite{langley00}

\bibliography{example_paper}
\bibliographystyle{icml2021}



\end{document}